\pdfoutput=1

\documentclass[11pt]{article}

\usepackage[preprint]{acl}
\usepackage{booktabs}
\usepackage{times}
\usepackage{latexsym}

\usepackage[T1]{fontenc}
\usepackage{amsmath}
\usepackage[utf8]{inputenc}

\usepackage{microtype}

\usepackage{inconsolata}

\usepackage{graphicx}

\usepackage{tcolorbox}
\usepackage{pifont}       
\usepackage{bbding}       
\usepackage{fontawesome}
\usepackage{soul}
\usepackage{float}
\usepackage{tabularray}
\usepackage{CJK}

\usepackage{diagbox}
\usepackage{enumerate}

\title{Moral Reasoning Across Languages: The Critical Role of Low-Resource Languages in LLMs}

\author{%
Huichi~Zhou\textsuperscript{1},
Zehao~Xu\textsuperscript{2},
Munan~Zhao\textsuperscript{2},
Kaihong~Li\textsuperscript{3},
Yiqiang~Li,
Hongtao~Wang\textsuperscript{2},
\\[0.6ex]              
\textsuperscript{1}Imperial College London \quad
\textsuperscript{2}North China Electric Power University \quad
\textsuperscript{3}Sun Yat-sen University}

\begin{document}
\maketitle

\begin{abstract}

In this paper, we introduce the \textit{Multilingual Moral Reasoning Benchmark} (MMRB)~\footnote{\url{https://anonymous.4open.science/r/ethics_scenarios-913B}} to evaluate the moral reasoning abilities of large language models (LLMs) across five typologically diverse languages and three levels of contextual complexity: sentence, paragraph, and document. Our results show moral reasoning performance degrades with increasing context complexity, particularly for low-resource languages such as Vietnamese. We further fine-tune the open-source LLaMA-3-8B model using curated monolingual data for alignment and poisoning. Surprisingly, low-resource languages have a stronger impact on multilingual reasoning than high-resource ones, highlighting their critical role in multilingual NLP.


\end{abstract}

\section{Introduction}

The advancement of Large Language Models (LLMs), such as ChatGPT~\citep{GPT-4}, has garnered significant attention in various downstream applications. Recently, many researchers have focused on the moral reasoning capabilities of LLMs and on preventing them from generating toxic or harmful content~\citep{zhou2023rethinking, rao2023ethical}. Unlike traditional language models (\emph{e.g.}, BERT~\citep{devlin-etal-2019-bert}), which tend to make binary moral judgments and are often criticized for their lack of interpretability~\citep{rao2023ethical, talat2022machine}, LLMs are capable of performing inference from a reasoning perspective (\emph{e.g.}, chain-of-thought~\citep{wei2023chainofthought}).

However, prior research on moral reasoning has mostly focused on monolingual datasets (primarily English), overlooking the importance of evaluating LLMs’ multilingual moral reasoning capabilities. The lack of robustness in multilingual moral reasoning may undermine the fairness of LLM-based systems, potentially affecting user experiences across different language backgrounds. Therefore, it is crucial to assess both the consistency and the key influencing factors of LLM performance in multilingual moral reasoning.

\vspace{5pt}
\textit{Do LLMs behave consistently on moral reasoning across different languages?}
\vspace{5pt}

To answer this question, we propose \textit{MMRB}, the \textbf{Multilingual Moral Reasoning Benchmark}, spanning five languages. Specifically, we collect well-established moral reasoning datasets, including ETHICS\textsubscript{BASE}~\citep{hendrycks2020aligning}, ETHICS\textsubscript{PRO}~\citep{ma2023oops}, and curate ETHICS\textsubscript{MAX} using the Moral Stories dataset~\citep{scherrer2024evaluating} and the moral dilemma framework~\citep{rao-etal-2023-ethical} to create more complex moral reasoning scenarios. These datasets are then translated into five typologically diverse languages (English, Chinese, Russian, Vietnamese, and Indonesian).

Based on \textit{MMRB}, we evaluate the multilingual moral reasoning performance of five prominent LLMs, including both closed-source (e.g., GPT-4~\citep{GPT-4}) and open-source (e.g., LLaMA~\citep{llama3modelcard}) models. Our experiments reveal significant inconsistencies in moral reasoning across different languages. To further explore the impact of language on model performance, we fine-tune LLaMA-3-8B on \textit{MMRB} to analyze cross-linguistic effects in alignment and poisoning settings. 

Our contributions are: (1) We introduce \textit{MMRB}, a multilingual moral reasoning dataset comprising 2,170 moral scenarios across five languages. (2) We benchmark five popular LLMs using \textit{MMRB}, revealing that their moral reasoning performance is inconsistent across languages. (3) We demonstrate that fine-tuning on low-resource languages has a disproportionately strong impact (both positive and negative) on multilingual reasoning performance.


\section{Related Work}
\paragraph{Monolingual Moral Reasoning.}

Most prior research on moral reasoning focuses on English. \citet{hendrycks2020aligning} proposed the ETHICS dataset, which covers concepts such as justice, well-being, duties, virtues, and commonsense morality, and evaluated models like BERT~\citep{devlin-etal-2019-bert}. \citet{ziems2022moral} introduced the MIC dataset to study whether chatbots can generate utterances aligned with Rules of Thumb (RoT) in moral contexts. The Moral Stories dataset~\citep{emelin-etal-2021-moral} examines whether language models can produce moral responses in social scenarios. To analyze LLMs' moral reasoning in ambiguous cases, \citet{scherrer2024evaluating} created datasets with 680 high-ambiguity and 687 low-ambiguity scenarios. However, these methods largely do not leverage LLMs' advanced reasoning capabilities, such as chain-of-thought (CoT) prompting~\citep{wei2022chain}, which can improve the stability and accuracy of moral inference. Moreover, they often overlook the generative nature of LLMs, which allows for explicit reasoning steps and nuanced decision-making beyond binary classification.

\paragraph{Multilingual Moral Reasoning.} According to the LLaMA-3 technical report,\footnote{\url{https://ai.meta.com/blog/meta-llama-3/}} LLMs are trained on different quantities of data per language, with over 5\% of the training data comprising high-quality non-English content covering more than 30 languages. As a result, performance is typically better in English. \citet{haemmerl-etal-2023-speaking} trained monolingual Sentence-BERT models for five languages to assess multilingual moral reasoning, but their approach focused on short, unambiguous sentences and high-resource languages. \citet{khandelwal2024moral} used GPT-4, ChatGPT, and LLaMA2-70B-Chat to evaluate moral dilemmas across six languages, finding weaker performance in Hindi and Swahili compared to Spanish, Russian, Chinese, and English. 

\section{Experiment}
\subsection{Setup}

\paragraph{Dataset.} 
We conduct multilingual alignment experiments using three datasets: (1) \textbf{ETHICS\textsubscript{BASE}}: A sentence-level binary classification dataset covering five moral dimensions (e.g., commonsense, justice). (2) \textbf{ETHICS\textsubscript{PRO}}: A paragraph-level dataset enhanced from ETHICS\textsubscript{BASE} using LLM-generated moral reasoning (e.g., ChatGPT), incorporating richer context and implicit ethical cues. (3) \textbf{ETHICS\textsubscript{MAX}}: A document-level dataset created by expanding paragraph-level entries into structured moral dilemmas. Each scenario involves three ethical branches (Virtue, Deontological, Consequentialist), each tied to six normative principles and three answer choices.

To support multilingual evaluation, we translate all datasets into five typologically diverse languages: English, Chinese, Russian, Vietnamese, and Indonesian. We use the DeepSeek API for initial translation, followed by manual verification by bilingual annotators. For quality assurance, we apply cross-validation using Google Translate to identify and correct translation inconsistencies or cultural mismatches. We also filter out scenarios with conflicting or ambiguous moral content, ensuring linguistic and conceptual fidelity. The resulting corpus, \textit{MMRB}, contains 2,170 high-quality scenarios across three complexity levels.

\paragraph{Model Selection.}
We evaluate five strong multilingual LLMs: GPT-4\footnote{\scriptsize \url{platform.openai.com/docs/models/gpt-4-turbo-and-gpt-4}}, GPT-3.5\footnote{\scriptsize \url{platform.openai.com/docs/models/gpt-3-5-turbo}}, LLaMA3-70B-Instruct\footnote{\scriptsize \url{replicate.com/meta/meta-llama-3-70b-instruct/api}}, LLaMA3-8B-Instruct\footnote{\scriptsize \url{replicate.com/meta/meta-llama-3-8b-instruct/api}}, and Mixtral-8x7B-Instruct-v0.1\footnote{\scriptsize \url{replicate.com/mistralai/mixtral-8x7b-instruct-v0.1/api}}. Inference is conducted using greedy decoding (temperature = $0$) for consistency.

\paragraph{Fine-tuning Strategy.}
To investigate how monolingual data quality affects multilingual performance, we fine-tune the open-source LLaMA3-8B model~\citep{llama3modelcard} using LLaMA-Factory~\citep{zheng2024llamafactory}. We design two fine-tuning settings: \textbf{Monolingual Alignment:} Fine-tuning on verified, high-quality data in a single language (e.g., Indonesian) to test cross-lingual generalization. \textbf{Monolingual Poisoning:} Fine-tuning on label-corrupted data in one language to examine its negative impact on other languages.

\subsection{Results}

\subsubsection{Evaluation}

\begin{table}[t]
\centering
\huge
\resizebox{\linewidth}{!}{
\begin{tabular}{l|ccccc} 
\toprule[2pt]
\diagbox{Dataset}{Model} & GPT-4 & GPT-3.5 & Llama3-70b & Llama3-8b & Mixtral  \\ 
\toprule
ETHICS\textsubscript{BASE-EN}         & 96.5 & 86.5   & 92.9       & 88.3     & 90.8          \\
ETHICS\textsubscript{BASE-ZH}         & 88.1 & 78.4   & 84.3      & 79.7      & 81.2          \\
ETHICS\textsubscript{BASE-RU}       & 91.7 & 75.3   & 84.9      & 76.2      & 84.2          \\
ETHICS\textsubscript{BASE-VI}         & 86.7 & 67.3   & 77.1      & 65.2     & 73.4         \\
ETHICS\textsubscript{BASE-ID}         & 90.3 & 75.6   & 79.2      & 63.0     & 72.5          \\
ETHICS\textsubscript{BASE-AVG}        & 90.6 & 76.6   & 83.7      & 74.5     & 80.4          \\
\hline
ETHICS\textsubscript{PRO-EN}          & 84.8 & 72.0    & 75.7       & 72.1     & 79.6          \\
ETHICS\textsubscript{PRO-ZH}          & 81.8 & 65.9   & 72.5      & 69.0        & 71.9          \\
ETHICS\textsubscript{PRO-RU}          & 82.9 & 64.5   & 74.7      & 64.6      & 74.5          \\
ETHICS\textsubscript{PRO-VI}          & 80.6 & 66.3   & 71.5      & 58.2      & 68.7          \\
ETHICS\textsubscript{PRO-ID}          & 82.9 & 68.9   & 67.1       & 57.1      & 71.1          \\
ETHICS\textsubscript{PRO-AVG}        & 82.6 & 67.5   & 72.3      & 64.2     & 73.2          \\
\hline
ETHICS\textsubscript{MAX-EN}          &  91.1	 & 78.5 &	87.4	& 84.0	& 63.7          \\
ETHICS\textsubscript{MAX-ZH}          & 90.6	&53.4	&77.7	&63.4	&66.4         \\
ETHICS\textsubscript{MAX-RU}          & 91.3	&55.4	&86.4	&78.0	&47.3        \\
ETHICS\textsubscript{MAX-VI}          & 85.6	&49.0	&80.7	&72.6	&27.6         \\
ETHICS\textsubscript{MAX-ID}          & 82.0	&50.1	&75.5	&70.2	&44.0         \\
ETHICS\textsubscript{MAX-AVG}        & 88.1	&57.3	&81.5	&73.7	&49.8         \\
\bottomrule[2pt]
\end{tabular}}
\caption{Evaluating the ability of moral reasoning across five LLMs on three datasets in multilingual setting.}
\label{tab:tab1}
\end{table}

Our goal is to evaluate the consistency of moral reasoning in multilingual settings across five LLMs on three levels of context: sentence, paragraph, and document. Table~\ref{tab:tab1} summarizes the results. We draw four main observations. (1) Even state-of-the-art models such as the GPT series do not produce consistent moral judgments across languages. We conduct Mann-Whitney U tests to measure statistical significance; most comparisons across languages fail to reach significance, indicating inconsistency in multilingual moral reasoning. (2) LLMs generally perform better in high-resource languages. For example, in ETHICS\textsubscript{BASE}, English accuracy ranges from $86.48\%$ to $96.49\%$, while Vietnamese (a low-resource language) lags behind, with scores ranging from $54.25\%$ to $86.71\%$. (3) Reasoning performance declines as context becomes more complex—from ETHICS\textsubscript{BASE} (sentence-level) to ETHICS\textsubscript{PRO} (paragraph-level). Interestingly, performance improves again on ETHICS\textsubscript{MAX} (document-level), likely because longer contexts include explicit moral principles that help guide the model’s decisions. This suggests that ambiguity in short prompts may confuse models, whereas explicit moral structure restores performance. (4) In Table~\ref{tab: tab4}, performance across the three ethical branches (Virtue, Deontological, Consequentialist) in ETHICS\textsubscript{MAX} varies, but follows similar trends across models and languages.

\begin{figure*}[ht]
    \centering
    \includegraphics[width=\textwidth]{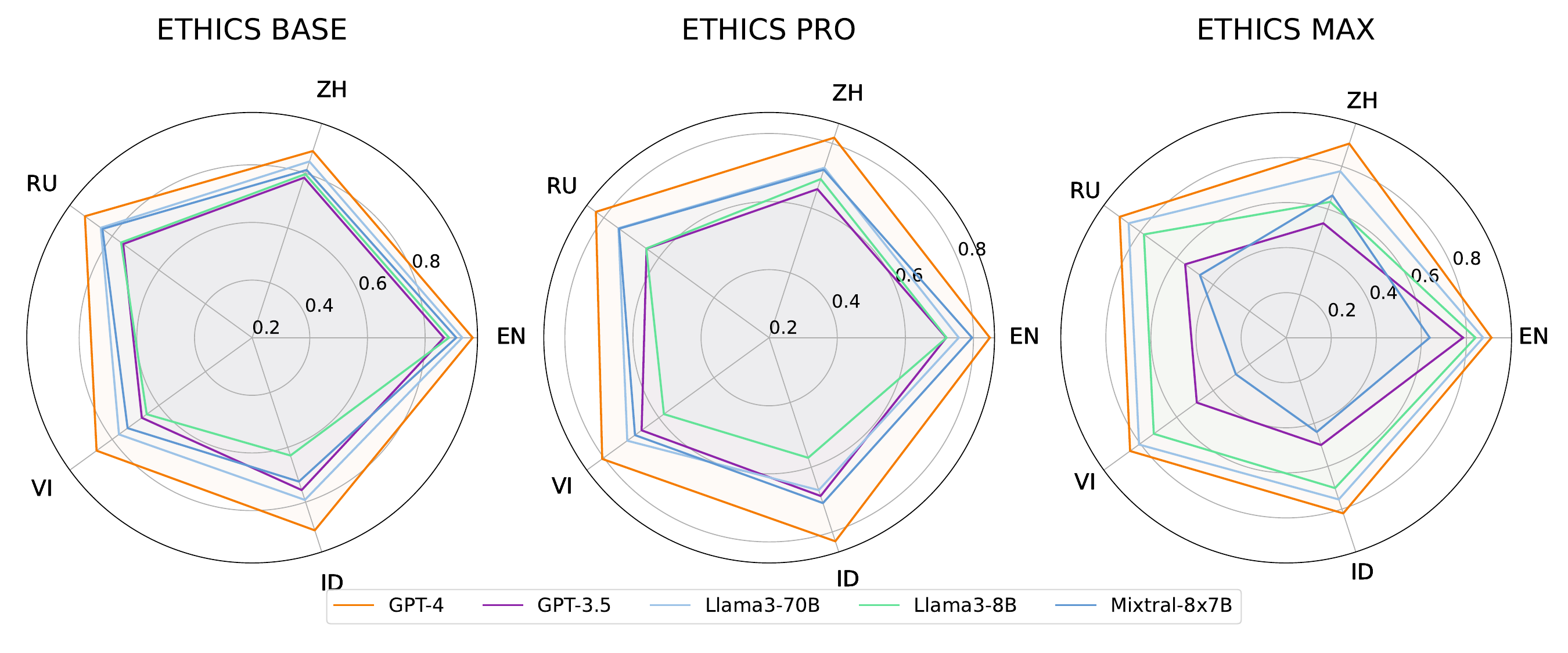}
     \caption{The performance of three levels datasets across five LLMs in a multilingual setting.}
    \label{fig:radar}
\end{figure*}

\begin{table}[t]
\huge
\centering
\resizebox{\linewidth}{!}{
\begin{tabular}{c|ccc|ccc|ccc|ccc|ccc}
\toprule[2pt]
   & \multicolumn{3}{c|}{GPT-4} & \multicolumn{3}{c|}{GPT-3.5} & \multicolumn{3}{c|}{Llama3-70b} & \multicolumn{3}{c|}{Llama3-8b} & \multicolumn{3}{c}{Mixtral} \\
   & V.  & D.  & C.  & V.  & D.  & C.  & V.  & D.  & C.  & V.  & D.  & C.  & V.  & D.  & C.  \\
\hline
EN & 94.5 & 91.0 & 89.2 & 79.0 & 82.9 & 76.1 & 89.1 & 89.5 & 85.7 & 85.5 & 85.9 & 82.1 & 63.4 & 66.1 & 61.5 \\
ZH & 94.5 & 91.8 & 89.3 & 51.3 & 53.8 & 52.9 & 77.4 & 81.1 & 76.2 & 62.5 & 65.7 & 62.0 & 68.6 & 67.4 & 63.8 \\
RU & 94.5 & 91.2 & 89.5 & 55.1 & 56.1 & 54.0 & 89.8 & 87.9 & 84.6 & 80.5 & 79.4 & 74.9 & 47.7 & 49.6 & 45.8 \\
VI & 89.9 & 86.7 & 83.1 & 47.8 & 48.5 & 47.9 & 82.1 & 83.5 & 77.9 & 73.5 & 75.0 & 69.9 & 30.0 & 27.5 & 28.2 \\
ID & 84.7 & 84.3 & 80.7 & 48.3 & 50.2 & 48.9 & 74.1 & 79.6 & 73.6 & 69.6 & 73.7 & 68.1 & 43.8 & 47.0 & 42.2 \\
\bottomrule[2pt]
\end{tabular}}
\caption{The moral reasoning accuracy rate of the LLMs across the three moral branches (Virtue, Deontological, Consequentialist).}
\label{tab: tab4}
\end{table}

\subsubsection{Monolingual Influence}

As LLMs are increasingly used across diverse regions and languages, ensuring consistent moral reasoning across linguistic boundaries becomes critical. However, low-resource languages often lack high quality and a large amount of data, which may impact the cross-lingual generalization of moral reasoning. To understand this phenomenon, we examine how fine-tuning a model on monolingual data (either clean or corrupted) affects its performance in other languages.

\begin{table}[t]
\centering
\huge
\resizebox{\linewidth}{!}{
\begin{tabular}{l|cccccc} 
\toprule[2pt]
\diagbox{Dataset}{Model} & Llama3\textsubscript{EN} & Llama3\textsubscript{ZH} & Llama3\textsubscript{RU} & Llama3\textsubscript{VI} & Llama3\textsubscript{ID}   \\ 
\toprule
ETHICS\textsubscript{BASE-EN}         & $\mathbf{94.2_{\uparrow 5.9}}$	&$\underline{91.9_{\uparrow 3.6}}$	& $91.2_{\uparrow 2.9}$	& $90.1_{\uparrow 1.8}$	&  $91.5_{\uparrow 3.2}$         \\
ETHICS\textsubscript{BASE-ZH}         & $81.5_{\uparrow 1.8}$	&$\mathbf{88.5_{\uparrow 8.8}}$	& $81.8_{\uparrow 2.1}$	& $82.9_{\uparrow 3.2}$	&$\underline{83.7_{\uparrow 4.0}}$   \\
ETHICS\textsubscript{BASE-RU}        & $80.4_{\uparrow 4.2}$ 	&$\underline{83.0_{\uparrow 6.8}}$	&$\mathbf{85.6_{\uparrow 9.4}}$	& $81.2_{\uparrow 5.0}$	& $79.9_{\uparrow 3.7}$       \\
ETHICS\textsubscript{BASE-VI}          & $65.5_{\uparrow 0.3}$	& $68.5_{\uparrow 3.3}$	& $80.3_{\uparrow 15.1}$	&$\mathbf{85.1_{\uparrow 19.9}}$	&$\underline{81.4_{\uparrow 27.1}}$           \\
ETHICS\textsubscript{BASE-ID}          & $70.6_{\uparrow 7.6}$	& $60.0_{\uparrow 1.5}$	& $\underline{78.5_{\uparrow 15.5}}$	& $78.1_{\uparrow 16.2}$	&$\mathbf{84.2_{\uparrow 21.2}}$       \\
\hline
ETHICS\textsubscript{PRO-EN}           & $\mathbf{84.6_{\uparrow 12.5}}$     & $82.9_{\uparrow 10.8}$      & $\mathbf{84.6_{\uparrow 12.5}}$      & $82.7_{\uparrow 10.6}$     & $\underline{84.5_{\uparrow 12.4}}$      \\
ETHICS\textsubscript{PRO-ZH}           & $72.2_{\uparrow 3.2}$      & $\mathbf{80.6_{\uparrow 11.6}}$      & $72.3_{\uparrow 3.3}$      & $75.9_{\uparrow 6.9}$      & $\underline{79.0_{\uparrow 10.0}}$         \\
ETHICS\textsubscript{PRO-RU}          & $76.4_{\uparrow 11.8}$      & $\mathbf{79.9_{\uparrow 15.3}}$     & $76.4_{\uparrow 11.8}$      & $75.7_{\uparrow 11.1}$      & $\underline{79.4_{\uparrow 14.8}}$           \\
ETHICS\textsubscript{PRO-VI}           & $64.4_{\uparrow 6.2}$     & $63.8_{\uparrow 5.6}$    & $64.4_{\uparrow 6.2}$      & $\mathbf{81.9_{\uparrow 23.7}}$      & $\underline{80.7_{\uparrow 22.5}}$           \\
ETHICS\textsubscript{PRO-ID}           & $61.5_{\uparrow 4.4}$      & $71.4_{\uparrow 14.3}$      & $61.5_{\uparrow 4.4}$      & $\underline{77.8_{\uparrow 20.7}}$     & $\mathbf{81.5_{\uparrow 24.4}}$        \\
\bottomrule[2pt]
\end{tabular}}
\caption{Monolingual alignment result.}
\label{tab: tab2}
\end{table}

\paragraph{Monolingual Alignment.} 

We fine-tune the model on high-quality, correctly labeled data from a single language, then evaluate performance on the remaining languages. As shown in Table~\ref{tab: tab2}, Indonesian fine-tuning yields the strongest improvements across other languages, followed by Vietnamese. These findings challenge the assumption that high-resource languages are always the most effective sources for cross-lingual transfer. On the contrary, low-resource languages may induce stronger alignment effects due to their underrepresentation in pretraining corpora, which allows fine-tuning to fill in critical knowledge gaps.

\begin{table}[t]
\centering 
\huge
\resizebox{\linewidth}{!}{
\begin{tabular}{l|cccccc} 
\toprule[2pt]
\diagbox{Dataset}{Model} & Llama3\textsubscript{EN} & Llama3\textsubscript{ZH} & Llama3\textsubscript{RU} & Llama3\textsubscript{VI} & Llama3\textsubscript{ID}   \\ 
\hline
ETHICS\textsubscript{BASE-EN}           & $\mathbf{80.2_{\downarrow 8.1}}$	&$85.2_{\downarrow 3.1}$	&$\underline{83.4_{\downarrow 4.9}}$	&$86.7_{\downarrow 1.6}$	&$86.1_{\downarrow 2.2}$            \\
ETHICS\textsubscript{BASE-ZH}          & $72.4_{\downarrow 7.3}$	&$\mathbf{68.0_{\downarrow 11.7}}$	&$\underline{72.0_{\downarrow 7.7}}$	&$78.4_{\downarrow 1.3}$	&$73.2_{\downarrow 6.5}$      \\
ETHICS\textsubscript{BASE-RU}            & $72.6_{\downarrow 3.6}$ & $\underline{70.9_{\downarrow 5.3}}$	&$\mathbf{67.6_{\downarrow 8.6}}$	&$76.1_{\downarrow 0.1}$	&$74.8_{\downarrow 1.4}$          \\
ETHICS\textsubscript{BASE-VI}   & $\underline{61.3_{\downarrow 3.9}}$	& $65.0_{\downarrow 0.2}$	& $61.7_{\downarrow 7.4}$	&$\mathbf{50.5_{\downarrow 14.7}}$	&$61.4_{\downarrow 3.8}$    \\
ETHICS\textsubscript{BASE-ID}            &  ${62.2_{\downarrow 0.8}}$         & $\underline{55.4_{\downarrow 7.6}}$        & $\mathbf{55.3_{\downarrow 7.7}}$          & ${58.1_{\downarrow 4.9}}$         & ${57.0_{\downarrow 6.0}}$           \\

\hline
ETHICS\textsubscript{PRO-EN}         & $\mathbf{62.90_{\downarrow 9.22}}$     & $68.30_{\downarrow 3.82}$      & $68.04_{\downarrow 4.08}$    & $\underline{66.63_{\downarrow 5.49}}$     & $71.18_{\downarrow 0.94}$        \\
ETHICS\textsubscript{PRO-ZH}         & $62.80_{\downarrow 6.20}$     & $\mathbf{59.02_{\downarrow 9.98}}$     & $\underline{59.40_{\downarrow 9.60}}$      & $63.70_{\downarrow 5.30}$     & $62.80_{\downarrow 6.20}$         \\
ETHICS\textsubscript{PRO-RU}           & $58.92_{\downarrow 5.68}$     & $58.80_{\downarrow 5.80}$     & $\underline{58.66_{\downarrow 5.94}}$      & $\mathbf{57.90_{\downarrow 6.10}}$     & $62.50_{\downarrow 2.10}$        \\
ETHICS\textsubscript{PRO-VI}          & $56.96_{\downarrow 1.24}$     & $53.29_{\downarrow 4.91}$     & $49.50_{\downarrow 8.70}$     & $\underline{39.50_{\downarrow 18.70}}$      & $\mathbf{35.90_{\downarrow 22.30}}$        \\
ETHICS\textsubscript{PRO-ID}        & $52.90_{\downarrow 4.20}$     & $50.85_{\downarrow 6.25}$     & $46.80_{\downarrow 10.30}$     & $\underline{45.10_{\downarrow 12.00}}$      & $\mathbf{28.40_{\downarrow 28.70}}$         \\
\bottomrule[2pt]
\end{tabular}}
\caption{Monolingual poisoning result.}
\label{tab: tab3}
\end{table}

\paragraph{Monolingual Poisoning.}

We further test the sensitivity of LLMs to data quality by fine-tuning the model on corrupted labels in one language, then observing performance in others. Table~\ref{tab: tab3} shows that poisoning with Vietnamese data causes the most degradation in multilingual performance, while English poisoning has minimal effect. This result suggests that low-resource languages are not only impactful in positive transfer (alignment) but also particularly vulnerable to negative influence, likely because harmful data in these languages is less diluted during pretraining. These observations underscore the importance of quality control in low-resource language data used for training.
\section{Discussion}

\begin{figure}[t]
    \centering
    \includegraphics[width=1.0\linewidth]{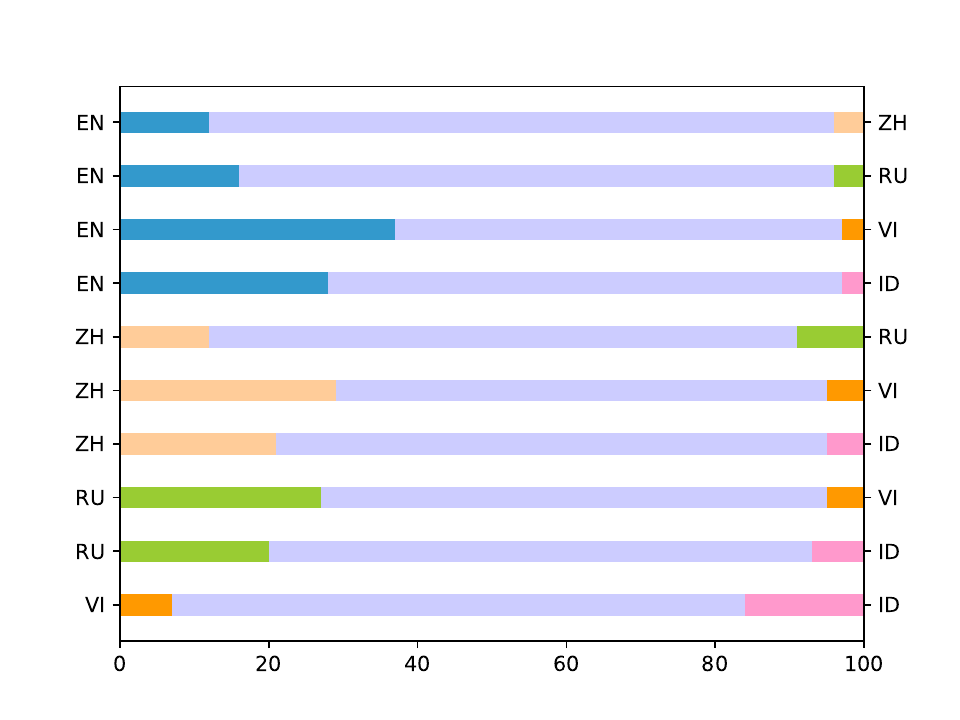}
     \caption{The win-rate between five languages in ETHICS\textsubscript{PRO} dataset based on Llama-3-8B.}
    \label{fig:win-rate}
\end{figure}

\paragraph{Win-Rate Analysis.} 
Beyond overall accuracy, we also examine pairwise win-rates between languages, as shown in Figure~\ref{fig:win-rate}. This metric captures subtle differences in model behavior across languages. Notably, while English benefits from extensive pretraining data, it still fails to generalize consistently in some moral reasoning cases, suggesting that training data volume alone does not guarantee robust moral understanding.

\paragraph{Model-wise Performance.}
Figure~\ref{fig:radar} illustrates the performance of all five LLMs across three datasets and five languages. GPT-4 consistently achieves the highest scores across all settings. In contrast, LLaMA-3-8B struggles particularly with sentence- and paragraph-level scenarios lacking explicit ethical principles. Its performance improves markedly on ETHICS\textsubscript{MAX}, where structured moral guidance is provided. Mistral-8x7B and GPT-3.5 show comparable performance on complex document-level tasks, indicating structured ethical inputs can mitigate some model limitations.

\paragraph{The Role of Low-Resource Languages.}
In Table~\ref{tab: tab2} and Table~\ref{tab: tab3}, low-resource languages play a pivotal role in shaping multilingual model behavior. On the one hand, high-quality data in these languages can significantly enhance generalization (alignment); on the other, corrupted or noisy data can severely impair moral reasoning across languages (poisoning). These findings emphasize the need for greater attention to data curation in under-resourced languages.


\section{Conclusion}

We introduce a multilingual benchmark for evaluating moral reasoning in LLMs. Our results reveal cross-lingual inconsistencies and show that low-resource languages have a strong impact (both positive and negative) on multilingual performance. These insights highlight the importance of data quality in underrepresented languages.

\section*{Limitations}
Despite the advances in assessing moral reasoning capabilities of LLMs across multiple languages, our study has limitations, including potential translation inaccuracies and model-specific performance variability. Furthermore, the scope of moral scenarios tested may not fully encompass real-world complexities, and additional research is needed to generalize our findings across diverse contexts.

\section*{Ethics Statement}
The ethical implications of fine-tuning LLMs, particularly regarding potential biases and fairness issues, require careful consideration. We are committed to transparency, accountability, respect for cultural diversity, and continuous improvement to ensure the development and deployment of responsible and ethical AI systems.

\label{sec:bibtex}

\bibliography{custom}

\appendix


\end{document}